\def\year{2022}\relax
\documentclass[letterpaper]{article} 
\usepackage{aaai22}  
\usepackage{times}  
\usepackage{helvet}  
\usepackage{courier}  
\usepackage[hyphens]{url}  
\usepackage{graphicx} 
\usepackage{natbib}  
\usepackage{caption} 
\DeclareCaptionStyle{ruled}{labelfont=normalfont,labelsep=colon,strut=off} 
\usepackage{mathrsfs}
\usepackage{multirow}
\usepackage{tabularx}
\usepackage{amsmath}
\usepackage{amssymb}
\usepackage{mmstyles}
\usepackage{color}
\usepackage[colorlinks,linkcolor=blue]{hyperref}

\title{DanceFormer: Music Conditioned 3D Dance Generation with Parametric Motion Transformer}

\author {
Buyu Li,\textsuperscript{\rm 1}
Yongchi Zhao,\textsuperscript{\rm 1}
Zhelun Shi,\textsuperscript{\rm 2}
Lu Sheng\textsuperscript{\rm 2}\thanks{Lu Sheng is the corresponding author.}
}
\affiliations {
    \textsuperscript{\rm 1} Huiye Technology,~~
    \textsuperscript{\rm 2} College of Software, Beihang University\\
    \{libuyu, zhaoyongchi\}@huiye.tech, ~ 
    \{18373044, lsheng\}@buaa.edu.cn
}

\begin{document}

\maketitle

\begin{abstract}

Generating 3D dances from music is an emerged research task that benefits a lot of applications in vision and graphics. Previous works treat this task as sequence generation, however, it is challenging to render a music-aligned long-term sequence with high kinematic complexity and coherent movements. In this paper, we reformulate it by a two-stage process, \ie, a key pose generation and then an in-between parametric motion curve prediction, where the key poses are easier to be synchronized with the music beats and the parametric curves can be efficiently regressed to render fluent rhythm-aligned movements. We named the proposed method as DanceFormer, which includes two cascading kinematics-enhanced transformer-guided networks (called DanTrans) that tackle each stage, respectively. Furthermore, we propose a large-scale music conditioned 3D dance dataset, called PhantomDance, that is accurately labeled by experienced animators rather than reconstruction or motion capture. This dataset also encodes dances as key poses and parametric motion curves apart from pose sequences, thus benefiting the training of our DanceFormer. Extensive experiments demonstrate that the proposed method, even trained by existing datasets, can generate fluent, performative, and music-matched 3D dances that surpass previous works quantitatively and qualitatively. Moreover, the proposed DanceFormer, together with the PhantomDance dataset (\url{https://github.com/libuyu/PhantomDanceDataset}), are seamlessly compatible with industrial animation software, thus facilitating the adaptation for various downstream applications.

\end{abstract}

\section{Introduction}

Automatically generating music conditioned 3D dances is an appealing but challenging task emerged in the research community of vision and graphics, which can significantly benefit various downstream applications in AR/VR, games, films and even the social networks.
Beyond human actions like walking, jumping and sitting with atomic functionalities, dancing is a type of artistic performances that focuses more on its choreography, \ie, sequential steps and movements with high kinematic complexity that are synchronized with the beats and rhythms of the music.
It is challenging for humans with professional training to generate expressive choreographies, no matter how hard a machine is to generate visually plausible dances accompanied with the music.

Most of the prior works~\cite{lee2019dancing2music,li2021dance,lee2018listen,tang2018dance_melody,li2020dance_transformer,zhuang2020music2dance,alemi2017groovenet} formulated the music conditioned 3D dance generation as a sequence generation problem, where each frame in the sequence describes the human pose by the joint-level rotations and translations.
However, rendering a long-term pose sequence with high kinematic complexity and coherent movements is still an open question.

In this paper, we would like to exploit a popular animation strategy~\cite{lasseter1987principles,thomas1995illusion} in the field of computer graphics, where the motion of characters is efficiently rendered by interpolating poses in keyframes through parametric curves.
This coarse-to-fine strategy is especially beneficial in our task since it both ensures kinematic complexity and motion conherency in the generated dances, where the poses in the keyframes can be generated and synchronized with the music beats due to their co-occurrence~\cite{lee2019dancing2music,li2021dance}, and the parameteric curves can be efficiently regressed so as to produce diverse motion patterns that are consistent with the rhythms of the music.
It also allows user-controlled temporal resolutions, and generates smooth sequences with fewer temporal flickers~\cite{williams2012animator}.
Moreover, formulating dances as a sequence of key poses and parametric motion curves would be seamlessly compatible with industrial animation pipelines.
Therefore, we would like to tackle the music conditioned 3D dance generation from this new perspective, and decompose this task into two easier sub-tasks, namely, the key pose generation and the subsequent parametric motion curve regression.

To this end, we propose DanceFormer, a two-stage music conditioned 3D dance generation framework that consists of two cascaded Transformer-like~\cite{vaswani2017attention} networks (called DanTrans), where the first one is to generate key poses and the second one is to regress the parameters of the motion curves between adjacent key poses.
At first, we employ an off-the-shelf beat track algorithm~\cite{ellis2007beat} to detect beats from the input music (\eg, around $30$ seconds).
The DanTrans for the key pose generation takes a sequence of music spectrum features centered at each beat as the input, and then generate a sequence of key poses that synchronizes with the sequence of the music beats.
And the DanTrans for the parametric motion curve regression employs a different sequence of music spectrum features that are extracted between adjacent beats, and feeds the generated key poses as well to regress the sequence of the parameters for each motion curve.
In our implementation, we employ the Kochanek-Bartels splines \cite{kochanek1984interpolating} to model motion curves, which benfits for rendering various movements from a small set of control parameters.

Each DanTrans contains a \emph{Wave Encoder} as a stack of transformer encoders to extract audio features, and a \emph{Motion Decoder} constructed by a stack of transformer decoders to auto-regressively predict key poses or parameters of the motion curves. 
To be specific, the Motion Decoder contains a new  Kinematic Propagation Module (KPM) and a Structured Multi-head Attention Module (SMAM) to enhance the spatial correlation for the dance generation process based on the Kinematic topological relations of the human body. This motion decoder increases the physical significance and thus alleviates unreasonable predictions seldom appeared in real dances.
%
%
These DanTrans networks are adversarially trained in accompany with $\ell_2$ reconstruction losses.

In order to train the proposed DanceFormer, we also propose a new PhantomDance dataset that is produced by professional animators rather than reconstruction from 2D videos~\cite{lee2019dancing2music,li2021dance,lee2018listen} or by motion capture~\cite{zhuang2020music2dance,tang2018dance_melody}.
The reconstructed data can not ensure the accuracy due to the limitation of recent 3D reconstruction algorithms. The motion capture methods are more accurate but still suffer from noise, such as limb flickers and joint mismatches.
In contrast, our PhantomDance dataset is produced by a team of experienced animators instructed by professional dancers. The animated dances are encoded by poses from keyframes and motion curve parameters, which thus can produce more fluent and expressive human pose sequences that matches the input musics.

Extensive experiments are conducted on PhantomDance and AIST++ \cite{li2021dance}, demonstrating that DanceFormer attains state-of-the-art results and significantly surpasses other works both qualitatively and quantitatively.

Above all, the contributions are summarized as follows:

(1) A new perspective to model the music conditioned 3D dance generation as key pose generation and parametric motion curve regression, which follows the animation principles and simultaneously ensures kinematic complexity and motion coherency.
%

(2) A novel framework named DanceFormer that is consists of two Transformer-like networks called DanTrans, where the first network generates key poses and the second one regresses the parameters of the motion curves.
More importantly, the DanTrans networks calculates attentions in adaptive temporal ranges and explicitly enhances the kinematic correlation among the outputs.

%
(3) The PhantomDance dataset (\url{https://github.com/libuyu/PhantomDanceDataset}), as the first dance dataset crafted by professional animators instead of 3D reconstruction or motion capture, provides more smooth and expressive dances that are synchronized with music, and directly compatible with industrial animation pipelines.

\section{Related Works}

\subsubsection{3D Dance Synthesis}
3D dance synthesis has been addressed in a number of ways. The traditional methods were based on retrieval from the dance dataset~\cite{2010Retrieval,chao2004motion}.
Recently, deep learning models have been widely employed in this task, attaining visually pleasing results, such as those using convolutional neural networks~\cite{lee2018listen,zhuang2020music2dance}, recurrent neural networks~\cite{tang2018dance_melody}, or using variational auto-encoders~\cite{lee2019dancing2music} to enhance the diversity of the generated dances.
CSGN~\cite{yan2019csgn} generates the dance sequences by enhancing the skeleton-level relations with a graph neural network (GNN), but the proposed Kinematic Propagation Module (KPM) used in our DanceFormer adopts forward and inverse kinematic message passing to better involve the kinematic correlations.
ChoreoMaster~\cite{chen2021choreomaster} introduces a choreography-oriented choreomusical embedding framework, and use the embedding to retrieve dance snippets in the motion graph for the input music. Because it doesn't generate new dances, we don't compare our method with it.
As the Transformer~\cite{vaswani2017attention} achieves a great success in sequence-to-sequence generation tasks in NLPs, some of the most recent works on the dance generation also bring it into use.
\citet{li2020dance_transformer} presented a two-stream motion transformer generative model.
AI Choreographer~\cite{li2021dance} presents a cross-model transformer with future-$N$ supervision for auto-regressive motion prediction.
These works directly generate the sequence of human poses, while the proposed DanceFormer generates a series of key poses and accompanied parameters of the motion curves, which ensures necessary kinematic complexity, motion coherency, adaptive temporal resolutions in the generated dances, and moreover, compatible with the industrial animation software, such as Maya, Unity and \emph{etc}.

\subsubsection{3D Dance Dataset}
%
%
The widely used motion datasets, such as Human3.6M~\cite{ionescu2013human3} and AMASS~\cite{mahmood2019amass}, collected common actions like walking, running, jumping and sitting. However, these datasets are hard to adapt for the dance generation task due to the huge distribution gap between daily motions and the dance movements, as well as the absence of carefully aligned music-dance pairs.
In fact, high-quality 3D dance movements with synchronized musics are extremely hard to collect.
Many existing datasets are limited by quantity or quality of data. Some of them just use sequences of 2D keypoints to represent dances~\cite{lee2019dancing2music,lee2018listen}.
AIST++~\cite{li2021dance}, as the largest dataset up to date, presented a $5$-hours 3D dance set. But it gathered dances by reconstructing 3D poses from 2D multi-view videos, thus the accuracy of pose parameters may not guaranteed.
The other works use motion capture to build dataset~\cite{alemi2017groovenet,tang2018dance_melody,zhuang2020music2dance}, in which the pose reliability is better but misalignment between the dance-music pair is inevitable.
On the contrary, the proposed PhantomDance is the largest public 3D dance dataset up to date and its data quality is much higher than existing datasets thanks to careful labels from experienced animators with the help of professional dancers. The stored key poses and parametric motion curves also benefit for animations by industrial softwares.

\section{The PhantomDance Dataset} 

\subsection{Data Collection}

We collect $260$ popular dance videos from more than $13$ genres from over $100$ different subjects (dancers) on YouTube, NicoNico and Bilibili, which have $9.6$ hours in total.
Then a team of experienced animators produced the 3D dance animations, as sequences of key poses and parametric motion curves.
An expert dancer provided professional instructions to teach the animators, thus improve the holistic expressiveness and detail richness of the animated dances, as well as the synchronization with beats and rhythms of the accompanied musics.
This dataset is named PhantomDance, and we will make it publicly available to facilitate future research. 

\subsection{Data Formatting}

In our PhantomDance dataset, the dances can be animated in industrial animation software, such as Maya~\cite{derakhshani2012maya}, on a standard SMPL~\cite{loper2015smpl} character model with $24$ body joints.
We followed Human3.6M~\cite{ionescu2013human3} and provided a subject-invariant 3D human skeleton representation for all the $260$ motion files, with unified limb lengths.
%
%
%
The motion is parameterized as the parametric curves of root (the hip joint) position and the rotations of the $24$ skeleton joints.
For each sequence we take the root position on the first frame as the the origin of the 3D coordinate. And the rotations are expressed with quaternions. That is, there are $3 + 4 \times 24 = 99$ curves in each motion file.
Each parametric motion curve is represented as the Kochanek-Bartels Spline interpolated from labeled key poses synchronized with extracted music beats, where curve segment between adjacent two key poses can be controlled by a fixed set of parameters. 

We use the parametric motion curves to represent dance movements because (1) its formulation is analytical and can be densely sampled with various temporal resolutions, which is very useful for real-time rendering; (2) its shape is in nature continuous and thus consistent with the aesthetic principles when evaluating dances.


\subsection{In Comparison to Previous Datasets}

Table~\ref{table:dataset} shows the comparison between our PhantomDance dataset with the other public music-conditioned 3D Dance Datasets.
The collected dances in PhantomDance mainly cover 13 genres, ranging from Urban Dance to Chinese Classical Dance, with extra genres from animes or idol groups which are hard to classify.
The beats per minute (BPM) of these dances range from $75$ to $178$.
Apart from labeling the detailed 3D positions and rotations of each pose, our dataset also provides the ground-truth music beats and parameters of each motion curve, thus directly facilitates animation or model learning that is compatible with industrial software.



Besides the aspect of dataset scale, the PhantomDance dataset also has high quality.
Motion data in the other datasets are collected from optical motion capture system~\cite{alemi2017groovenet,tang2018dance_melody,sun2020deepdance} or 3D reconstruction from multi-view videos~\cite{li2021dance}.
Motion capture systems can not totally avoid data noise like limb flicker and joint mismatch.
And video based 3D reconstruction suffers from more serious noise like foot slippery and limb twist due to the limitation of reconstruction algorithms.
Thus professional animators are needed for data correction.
In contrast, motion data in PhantomDance are produced by expert animators from scratch with the guidance of a professional dancer. And we have a strict quality checking process so that it costs about $15$ months to finish the dance labeling.
More details of our PhantomDance dataset and the qualitative comparison with the other datasets can be seen in our project page\footnote{https://huiye-tech.github.io/post/danceformer/}.

\begin{figure*}[t]
\centering
\includegraphics[width=\textwidth]{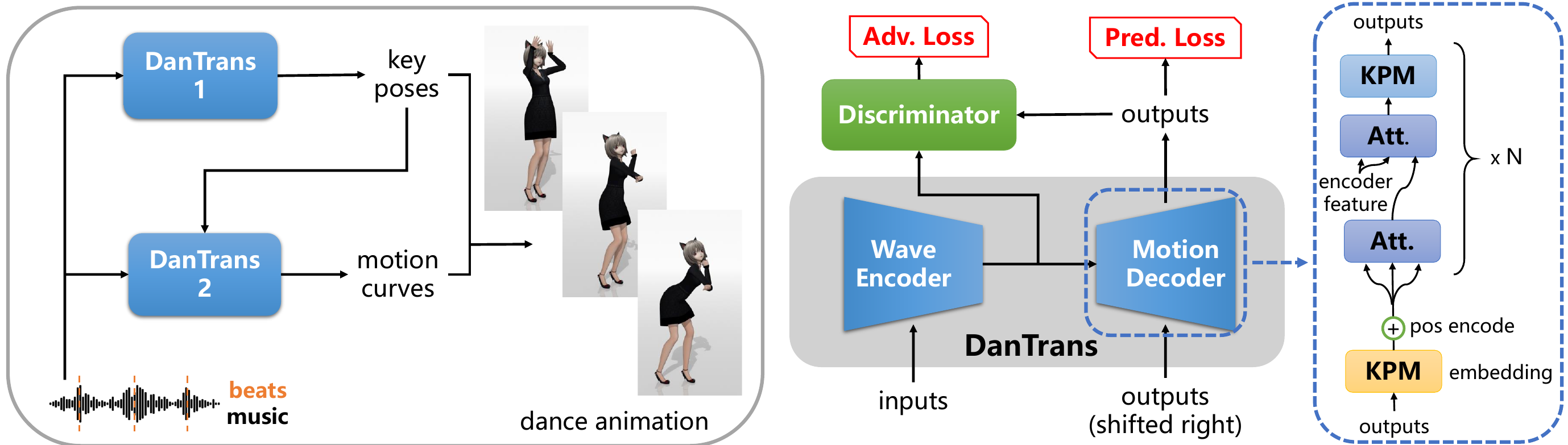}
\caption{{\bf Overview of the workflow of DanceFormer.} 
         The networks in stage 1 take as input the music wave features in the neighborhood of each beat and output a sequence of key poses. Then the networks in stage 2 utilize the generated key poses and the music wave features between each two key poses to predict the motion curve in between. Each network adopts the proposed transformer-like network, called DanTrans. The DanTrans has a new transformer decoder, which uses the proposed Kinematic Propagation Module (KPM). Note that the attention module (Att. in the figure) is a structured multi-head attention.
         }
\label{fig:overview}
\vspace{-3mm}
\end{figure*}

\begin{table}[t]
\centering
\resizebox{\linewidth}{!}{
     \begin{tabular}{ c | c c c c c c }
     \hline
     Dataset & Pos/Rot & Beat & Curve & Genres & Music & Seconds  \\
     \hline
     GrooveNet & $\checkmark$/$\times$ & $\times$ & $\times$ & 1 & 3 & 1380 \\
     EA-MUD & $\checkmark$/$\times$ & $\times$ & $\times$ & 6 & 23 & 1849 \\
     Dance w/ Melody & $\checkmark$/$\times$ & $\times$ & $\times$ & 4 & 61 & 5640 \\
     AIST++ & $\checkmark$/$\checkmark$ & $\times$ & $\times$ & 10 & 60 & 18694 \\
     PhantomDance & $\checkmark$/$\checkmark$ & $\checkmark$ & $\checkmark$ & 13+ & 260 & 34667 \\
     \hline
     \end{tabular}}
     \vspace{-2mm}
     \caption{Our PhantomDance dataset \emph{v.s.} the other datasets, namely GrooveNet~\cite{alemi2017groovenet}, EA-MUD~\cite{sun2020deepdance}, Dance with Melody~\cite{tang2018dance_melody} and AIST++~\cite{li2021dance}.}
     \label{table:dataset}
\end{table}

\section{DanceFormer}

As illustrated in Figure~\ref{fig:overview}, we propose the DanceFormer, a two-stage music conditioned 3D dance generation framework that consists of two cascaded transformer-like networks. Each network is named as DanTrans, where the first one generates key poses, and the second one predicts the parameters of the motion curves between adjacent key poses.
Each DanTrans network contains a wave encoder that is similar as the standard transformer encoder~\cite{vaswani2017attention}, and a motion decoder with a Kinematic Propagation Module (KPM) and a Structured Multi-head Attention Module (SMAM), which enhances the kinematic correlation for the dance generation process based on the Kinematic topological relations of the human body.
Adversarial learning is applied during the training in both stages.


\subsection{Stage 1: Key Pose Generation}

Following the rhythm (beats) is a basic principle in dance theory \cite{goodridge1999rhythm}, and previous studies also show that music beats and kinematic inflexions have strong consistency in time \cite{lee2019dancing2music,li2021dance}.
Thus we define the key poses directly as the poses on the beats, which is represented as position and rotation (in quaternion) of each skeleton joint, with the dimension of $(3 + 4) \times 24 = 168$.
We first leverage the a dynamic programming algorithm~\cite{ellis2007beat} to track the time of each beat in the music.
Our target in stage 1 is to generate a sequence of key poses to match these extracted beats.

The input sequential music features are music spectrum features between a window centered at each beat.
We use a window of size $0.8$ seconds in our implementation since the lowest beat per-minus (BPM) is $75$ in common dance musics (which is validated in our PhantomDance dataset, and also valid in other datasets).
We choose the Hamming window to tackle the window overlap in music with higher BPMs.
In each window, we calculate the $40$-dimensional form of MFCC~\cite{logan2000mfcc} that has $13$ coefficients of Mel filters cepstrum coefficients with the first and second order differences ($13 \times 2$ dimensions) and the energy sum (1 dimension).
We further append a $12$-dimension chroma feature to the feature of every beat and finally attain the inputs to the wave encoder.
And the motion decoder combines the encoded features with the right shifted outputs and sequentially predict the key poses, as indicated by the sequence-to-sequence prediction strategy based on the standard transformer~\cite{vaswani2017attention}.

%


\begin{figure}[t]
\centering
\includegraphics[width=\linewidth]{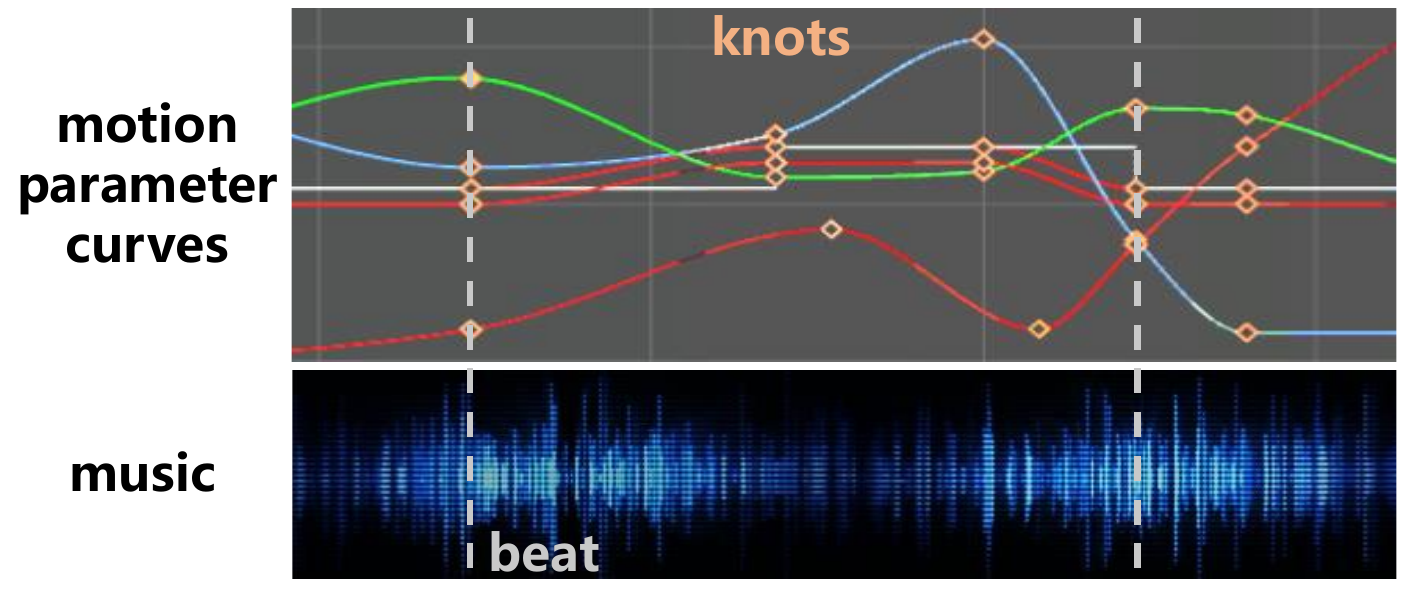}
\caption{{\bf Motion Curves.} 
         The figure shows an example of $4$-knot TCB spline parameterization for motion curves.}
\label{fig:curve}
\vspace{-3mm}
\end{figure}

\subsection{Stage 2: Parametric Motion Curve Prediction}

Based on the generated key poses, the DanTrans in stage 2 is targeted to generate the motion curves in between.

The motion curves are defined as the values of transformation parameters with regard to time. The transformation parameters are translation $(x, y, z)$ and rotation $(r_x, r_y, r_z, r_w)$ in quaternion of each joint.
We use a multi-knots Kochanek-Bartels splines~\cite{kochanek1984interpolating} to model each curve.
It is a sort of cubic Hermite spline that is used in animation editing softwares. 
The Kochanek-Bartels spline is also named TCB spline because it has $3$ parameters: $t$ for tension, $b$ for bias and $c$ for continuity. While for a multi-knot TCB spline, we should also determine the intermediate knots and their tangents.
Figure~\ref{fig:curve} shows a example of motion curve parameterization.


In our implementation, we use one 4-knot TCB spline to fit one motion curve between two adjacent key poses, as an optimal tradeoff between fitting accuracy and representation compactness.
Since the endpoints in the curve are just on the key poses, we have $7$ control parameters to predict, namely the $t$, $c$, $b$ and the position of the two intermediate knots.


Similar to the DanTrans in stage 1, we extract the sequence of the MFCC features from the music between every two adjacent beats as the input for the wave encoder, and the motion decoder auto-regressively outputs the $7$ control parameters that are aligned with each MFCC feature.
Note that the motion decoder requires the right shifted output sequences as its input, which should be additionally concatenated with the two key poses at the beginning and the end of the curve, which are generated in stage 1.

\subsection{Network Structure of DanTrans}

DanTrans has a similar Network structure to the standard transformer~\cite{vaswani2017attention} as illustrated on the right of Figure~\ref{fig:overview}, except that the motion decoder employs the proposed Kinematic Propagation Module (KPM) and the Structured Multi-head Attention Module (SMAM) in each transformer decoder layer.
We employ a stack of $N=6$ transformer decoders in the motion decoder of each DanTrans. 
Note that an additional KPM is applied at the beginning of the motion decoder for output embedding.

\subsubsection{Kinematic Propagation Module}

\begin{figure}[t]
\centering
\includegraphics[width=\linewidth]{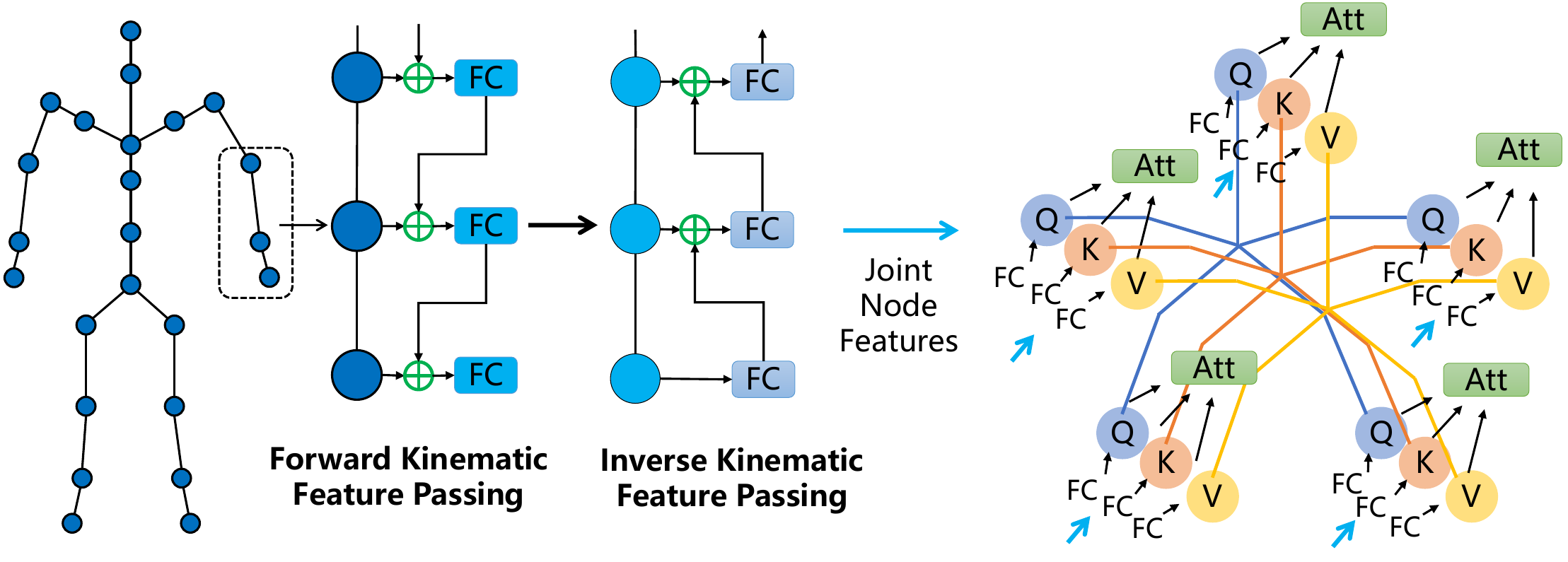}
\caption{{\bf The Kninematic Propagation Module (KPM) and the Structured Multi-head Attention.} 
         Features are embedded and passed along the kinematic chains forward and inversely with fusion operations. Note that the endpoints of feature passing are root and leaf nodes, and we just show a part of them for clearer view.}
\label{fig:KPM}
\vspace{-3mm}
\end{figure}

The Kinematic Propagation Module (KPM) are constructed on the basis of human body structure, \ie, the $24$ skeleton joint nodes with a tree topology. The networks consist of a stack of so-called FK-IK block as illustrated in Figure~\ref{fig:KPM}.
Note that the fully connected layer (FC) represents the complete BN-Linear-ReLU module, while we use FC to represent it for simplicity.

The FK-IK block has two steps, \ie, the forward kinematic (FK) feature passing and then the inverse kinematic (IK) feature passing, respectively. 
The forward kinematic feature passing step firstly performs a feature embedding on the root node and pass it to the neighbor child nodes along the kinematic chain.
After receiving the features from the parent node, the child node performs feature fusion and then conducts feature embedding by linear projection.
The FC layers are utilized for feature embedding and the fusion adopts the addition operation.
The backward kinematic feature passing step has a similar feature embedding and the passing process is performed from the leaf nodes to the root.
These procedures are similar to the operations of forward kinematics (FK) and inverse kinematics (IK) in computer graphics and robotics.
The FK-IK block attempts to encode features into the motion controlling parameter space and to bring physics constraints into the model.

For the output embedding process at the beginning of the motion decoder, we use a KPM with one FK-IK block. And for feed-forward process in the intermediate layers, we adopt a stack of 2 FK-IK blocks for each KPM.
All the KPMs have the same output feature size of $64$.

\subsubsection{Structured Multi-Head Attention}

The features outputted by KPM are arranged as the joint nodes, they naturally act as heads in the multi-head attention mechanism used in common transformers~\cite{vaswani2017attention}.
In the multi-head cross-attention, the encoded features from the input music are processed by two sets of 24 paralleled FC layers with the output size of $64$, which then act as the values and keys.
Note that no concatenation is needed after attention calculation since the features should maintain the body structure, and directly fed into the succeeding KPM module.
Thus this process saves computation compared to the typical multi-head attention that concatenates features from each head and uses a big linear layer for feature embedding.
The structured multi-head attention module, also shown in the right of Figure~\ref{fig:KPM}, defines the head as the node in the body structure, which has explicit physical significance. To some degree this design is more reasonable than the multi-heads in the original transformer.


\subsection{Training Objective}

We adopt an adversarial training strategy that regards each DanTrans network as an individual generator, and learns two discriminators to judge whether distributions of the generated sequences and the groundtruth ones are aligned or not.
Each discriminator consists of two modules to process the music feature and the output/groundtruth key pose sequence (in stage 1) or the parameters of the motion curves (in stage 2), and the processed features are combined and fed into the binary classifier to predict the authenticity of the outputs conditioned on the input music. All modules in the discriminator are 3-layer MLPs. 
Moreover, when updating the generator, we also include an $\ell_2$ reconstruction loss to avoid unwanted solutions that are far away with real data.

\section{Experiments}


\subsection{Implementation Details}

Since most musics in PhantomDance have the verse-chorus form, we divide the $260$ dance animations into $1000$ sequences.
Among them $900$ pieces of music-dance pairs are used for model training, and the other $100$ are split into the test set. We carefully pick the test set to ensure it cover $13$ genres and the BPM range from $80$ to $180$.
We follow the official training/testing splits in AIST++~\cite{li2021dance}. To gather the ground-truth labels for the training of the DanceFormer, we use the beat track algorithm~\cite{ellis2007beat} to extract beats and fit the parameter motion curves from the provided pose sequences.
The DanceFormer is end-to-end trained using $4$ TITAN Xp GPUs with a batch size of $8$ on each GPU.
We use the Adam optimizer with betas $\{0.5, 0.999\}$ and a learning rate of $0.0002$.
The learning rate drops to $2$e$-5$, $2$e$-6$ after {$100$\emph{k}, $200$\emph{k}} steps.
The model is trained with $300$\emph{k} steps for \emph{AIST++} and $400$\emph{k} steps for PhantomDance.
The dimension of features in DanceFormer is $256$ unless otherwise specified.

\subsection{Evaluation Metrics}

\subsubsection{Normalized Power Spectrum Simularity (NPSS)}

It is an evaluation metric for long-term motion synthesis compared to Mean Square Error (MSE). 
We just follow its official implementation \cite{gopalakrishnan2019neural_rnn} and compute NPSS in the joint motion space $\mathbb{R}^{T \times N \times 7}$ ($4$ for joint rotation represented as quaternion and $3$ for joint position).


\subsubsection{Frechet Distance (FD)} 

It is proposed by AIST++~\cite{li2021dance} that has two metrics, namely PFD for position and VFD for velocity. We also employ it to calculate the distribution distance of joints.

\subsubsection{Position Variance (PVar)}
It evaluates the diversity of the generated dance.
In AIST++, a piece of music corresponds to more than one dance, but only one in PhantomDance. So we make a modification to the metric~\textbf{PVar} in the experiments on PhantomDance where we compute it along different music pieces which have identical length.

\subsubsection{Beat Consistency Score (BC)}

It is a metric for motion-music correlation. We follow~\cite{li2021dance} to define kinematic beats as the local minima of the kinetic velocity.
BC computes the average distance between every music beat and its nearest kinematic beat with the following equation:
\begin{equation}
BC = \frac{1}{|\cB^x|} \sum_{i=1}^{|\cB^x|} \exp\left(-\frac{\min_{\forall t_j^x \in \cB^x}||t_j^x - t_i^y ||^2}{2\sigma^2}\right)
\end{equation}
where $\cB^x = \{ t^x_j \}, \cB^y = \{ t^y_i \}$ are kinematic beats and music beats respectively and $\sigma = 3$.
Note that BC has a similar form to Beat Alignment Score (BA) proposed in \cite{li2021dance}, but they are different in essence.
BA forces every kinematic beat to match a music beat, but a dance usually has many small kinematic beats that occur between music beats.
Moreover, a music synchronized dance just need to ensure the most salient music beats are accompanied with the action emphasis (kinematic beats). So our proposed BC, which finds matched kinematic beat for each music beat, is more appropriate in this case.

\begin{table}[t]
 \centering
 \resizebox{\linewidth}{!} {
     \begin{tabular}{ c | c  c  c  c  c }
     \hline
     Component & NPSS $\downarrow$ & PFD $\downarrow$ & VFD $\uparrow$ & PVar $\uparrow$ & BC $\uparrow$ \\
     \hline
      \textbf{DanceFormer} & \textbf{8.03} & \textbf{114.03} & \textbf{0.55} & \textbf{0.912} & \textbf{0.785} \\
     \hline
     Curve to Frames & 14.55 & 1132.4 & 1.06 & 0.233 & 0.317 \\
     \hline
     KPM to Linear & 10.54 & 162.44 & 0.79 & 0.592 & 0.637 \\
     KPM to GNN & 10.34 & 146.43 & 0.69 & 0.647 & 0.692 \\
     \hline
     \end{tabular}
 }
\vspace{-2mm}
     \caption{{\bf Ablation study about our DanceFormer on the PhantomDance dataset.} $\downarrow$ means that lower results indicate better methods, and $\uparrow$ \emph{vice versa}.}
     \label{table:ablation}
\end{table}


\subsection{Ablation Study}

To validate the effectiveness of our DanceFormer, we conduct the ablation study on our PhantomDance dataset.

\begin{table}[t]
 \centering
 \resizebox{\linewidth}{!} {
     \begin{tabular}{ c | c  c  c  c  c }
     \hline
     Att Algorithm & NPSS $\downarrow$ & PFD $\downarrow$ & VFD $\uparrow$ & PVar $\uparrow$ & BC $\uparrow$ \\
     \hline
     Global & 9.73 & 151.53 & 0.72 & 0.649 & 0.737 \\
     Local & 9.01 & 125.59 & 0.68 & 0.744 & 0.763 \\
     Gaussian Local & 8.93 & 117.14 & 0.64 & 0.753 & 0.764 \\
     \textbf{Learned Local} & \textbf{8.03} & \textbf{114.03} & \textbf{0.55} & \textbf{0.912} & \textbf{0.785} \\
     \hline
     \end{tabular}
 }
 \vspace{-2mm}
     \caption{{\bf The comparison of different Attention mechanisms.} $\downarrow$ means that lower results indicate better methods, and $\uparrow$ \emph{vice versa}.}
     \label{table:att}
      \vspace{-5mm}
\end{table}

\begin{figure*}[t]
\centering
\includegraphics[width=\textwidth]{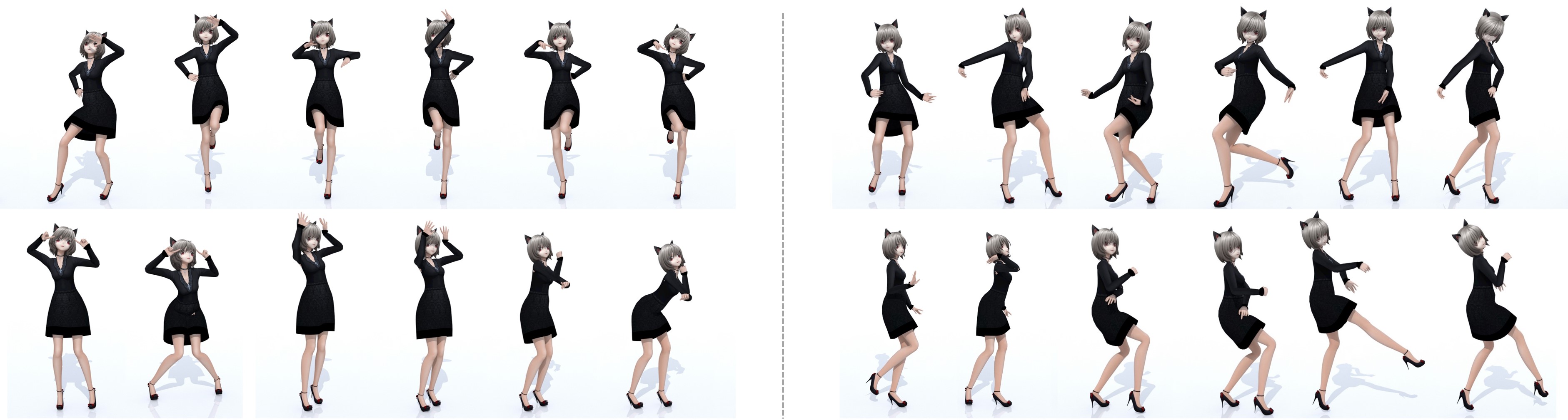}
\caption{{\bf Visualization of generated results of DanceFormer.}}
\label{fig:result}
\end{figure*}

\begin{figure}[t]
\centering
\includegraphics[width=\linewidth]{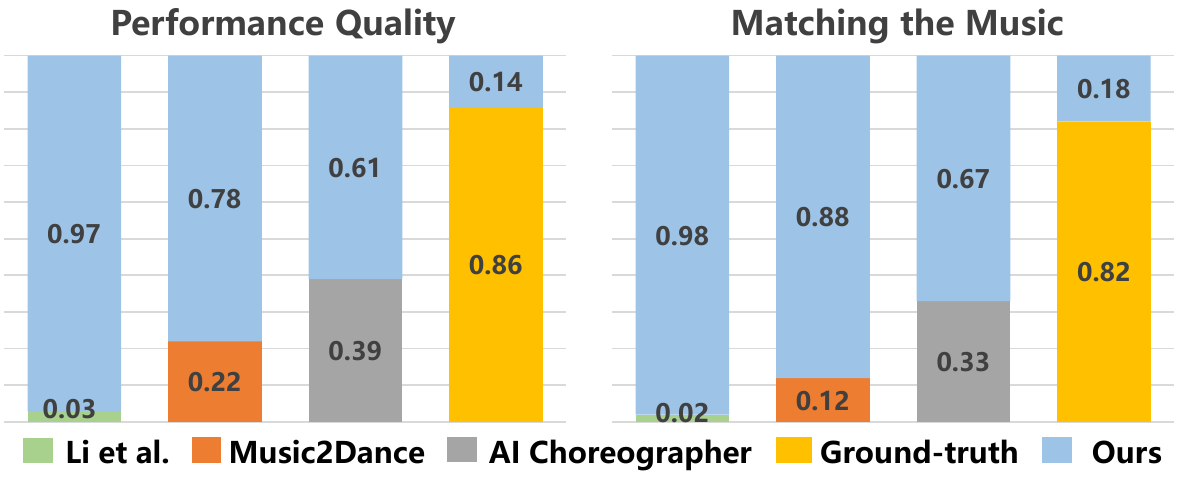}
\vspace{-3mm}
\caption{{\bf Results of the user study.}  We conduct a user study to ask participants to choose the better dances from pairwise comparisons. The criteria includes the performance quality and matching the music. The number denotes the percentage of preference on the comparison pairs.}
\label{fig:userstudy}
\vspace{-3mm}
\end{figure}

\begin{table*}[!t]
\centering
 \resizebox{\linewidth}{!}{\begin{tabular}{ c | c | c | c | c | c || c | c | c | c | c }
 \hline
 \multirow{2}{*}{Method} & \multicolumn{5}{c||}{AIST++} & \multicolumn{5}{c}{PhantomDance} \\
 \cline{2-11}
 & NPSS $\downarrow$ & PFD $\downarrow$ & VFD $\uparrow$ & PVar $\uparrow$ & BC $\uparrow$ & NPSS $\downarrow$ & PFD $\downarrow$ & VFD $\uparrow$ & PVar $\uparrow$ & BC $\uparrow$ \\
 \hline     
 Li et al. \cite{li2020dance_transformer} & 16.31 & 5595.91 & 3.40 & 0.019 & 0.359 & 18.34 & 7944.78 & 5.84 & 0.014 & 0.175 \\
 Music2Dance \cite{zhuang2020music2dance} & 14.74 & 2367.26 & 1.13 & 0.215 & 0.378 & 15.94 & 3147.89 & 3.74 & 0.267 & 0.223 \\
 AI Choreographer \cite{li2021dance} & 8.29 & 113.56 & 0.45 & 0.509 & 0.452 & 10.62 & 164.33 & 0.73 & 0.624 & 0.388 \\
 \hline
 DanceFormer & \textbf{6.01} & \textbf{84.32} & \textbf{0.34} & \textbf{0.734} & \textbf{0.782} & \textbf{8.03} & \textbf{114.03} & \textbf{0.55} & \textbf{0.912} & \textbf{0.785} \\
 \hline
 \end{tabular}}
 \vspace{-2mm}
     \caption{{\bf Dance generation evaluation on the AIST++ and the PhantomDance datasets.} Our method outperforms other baselines in terms of quality, diversity and beat consistency. Especially, due to our two-stage prediction schema, our model has superior performance in term of BC indicating that our model can generate dances which better match the given music. $\downarrow$ means that lower results indicate better methods, and $\uparrow$ \emph{vice versa}.}
    \label{table:comparison}
     \vspace{-2mm}
\end{table*}

\subsubsection{Effectiveness of Curve Prediction}
 
To evaluate the effectiveness of our curve prediction, we sample 60-FPS poses from the data and use a standard transformer to directly predict the sequence. The input sequences are extracted from a $0.5$s sliding window with a step of $1/60$ second on the music waves. Its result is the second line of Table~\ref{table:ablation} as ``Curve to Frame". The experimental result demonstrates the advantage of our pose-to-curve two stage generation framework.
 
 \subsubsection{Effectiveness of KPM}

 To study the impact of the KPM component, we first replace it with the standard feed-forward (linear layers) modules in vanilla transformer \cite{vaswani2017attention}. And the multi-head attention also comes back to the common form. The third line in Table~\ref{table:ablation} shows that the KPM has significant improvement in motion generation quality.
 To further study the FK-IK processing, we introduce a comparison architecture which substitutes the FK-IK block with a 2-layer graph neural network (GNN)~\cite{scarselli2008graph}.
 This GNN fuses the joint features according to the adjacency matrix of undirected graph defined by the joint structure.
 The experimental result shows that GNN has better performances compared to the baseline (``KPM to Linear''), which proves that introducing spatial correlations benefits the representation of the learning model.
 While the proposed KPM surpasses the trivial GNN by a large margin, validating that involving spatial correlation and physics constraints into the network brings in  significant improvements in the generation quality that is evaluated by NPSS, PFD and VFD.
 %

\subsubsection{Variants of Attention}

In our implementation, we find that local attention~\cite{luong2015effective} is more proper for the motion sequence generation task than the global attention in standard transformers.
It can be explained by the temporal locality of motions.
That is, the pose several seconds ago has no direct influence on the current pose.
Moreover relationship between motion state is stronger with the time going closer. So we further compare several attention algorithms and the results are shown in Table~\ref{table:att}. We use a sequence length of $17$ in the simple local attention experiment shown on the second line. And we add a Gaussian mask on the attention results before \texttt{softmax} with a standard variance of $4$. The result is on the third line and it has a slight improvement. Finally we use a learned mask to be optimized with the networks and obtain the best result. The trained mask has a triangle shape on the whole with some humps. Our DanceFormer employs this learnable local attention in our implementation.

 

\subsection{Comparison with Other Methods}

We mainly compare our method with AI Choreographer~\cite{li2021dance}, which to our knowledge obtains the state-of-the-art results for music-conditioned dance generation. The other two most related works, namely Li et al.~\cite{li2020dance_transformer} and Music2Dance~\cite{zhuang2020music2dance} are also compared.
The experimental comparisons are performed on both the AIST++ and our PhantomDance datasets.

\subsubsection{Quantitative Comparisons}

The results are viewed in Table~\ref{table:comparison}. Our method outperforms Li et al.~\cite{li2020dance_transformer} and Music2Dance~\cite{zhuang2020music2dance} by a considerably large margin. And it also surpasses AI Choreographer~\cite{li2021dance} significantly in the quality related metrics (a $30\%$ gain of NPSS, a $28\%$ gain of PFD and a $27\%$ gain of VFD), diversity metrics (a $46\%$ gain of PVar) and beat consistency metrics (a $93\%$ gain of BC).
The promising results on the dance quality metrics are mainly owing to the network structure of KPM that enhances the kinematic correlations. The generation diversity is ensured by the adversarial learning scheme.
Note that the upper bound of the metric BC is $1$, which means there exactly exists a kinematic beat at the time of each music beat. So our results have relatively high beat consistency score, which is due to the proposed two-stage framework of our DanceFormer.
Since no metric emphasizes the fluency of the dance performance, the advantages of motion curve formulation can only be well revealed in qualitative results, as shown in Figure~\ref{fig:result}.

\subsubsection{Qualitative Results and User Study}

Figure~\ref{fig:result} provides a sequence of frames about the generated dances by our DanceFormer. These results show that the proposed method can provide diverse movements with high kinematic complexity.
Video results accompanied with musics, including the comparisons with the other works can be found in the project page\footnote{https://huiye-tech.github.io/post/danceformer/}. 

We also conducted a user study to evaluate the quality of the generated music-conditioned dances.
All the $100$ sequences of the validation set of PhantomDance were used for the study.
And then we collect the generated dances by our method and the compared baselines~\cite{li2021dance,li2020dance_transformer,zhuang2020music2dance}.
In addition, the ground-truth dances are also included.
The user study was conducted using a pairwise comparison scheme.
For each of the $100$ musics, we provide $4$ pairs in which our results occur with the results from the baseline methods or the ground truth.
Thus $400$ pairs were provided to the participants, and they were asked to make two choices for each pair:
``Which dance is a better performance (more fluent, graceful and pleasing)?'' and
``Which dance matches the music better?''.
There are $100$ participants in the user study.
Figure~\ref{fig:userstudy} shows the user study results, where our DanceFormer outperforms the other methods on both criteria.
Most of the participants rated that our method generates better dances in performance quality compared with other works, and even more participants held the opinion that the dances generated by our model better match the musics.

\section{Conclusion}
In this work, we propose a new perspective to model the music-conditioned 3D dance generation task.
Different from previous works that define the outputs as sequences of poses, we formulate them as key poses and in-between motion curves.
The curve representation makes the generated results more fluent and graceful.
Based on this formulation, we propose the transformer-based DanceFormer with the novel DanTrans architecture consisting of the KPM module for better modeling kinematic correlations.
DanceFormer thus yields high-quality results in the experimental comparisons.
Moreover, we propose the PhantomDance Dataset, the first music-conditioned 3D dance dataset that uses curves to represent body motion, and it is the largest 3D dance dataset with the best visual quality up to date. 

\subsubsection{Acknowledgement}
This work was partially supported by the National Natural Science
Foundation of China (No. 61906012, No. 62132001).

\bibliography{aaai22}

\end{document}